\documentclass[table]{article} 
\usepackage{iclr2024_conference,times}

\usepackage{hyperref}
\usepackage{url}
\usepackage{tikz}
\usepackage{multirow}
\usepackage{xspace}
\usepackage{booktabs}
\usepackage{ulem}
\usepackage{graphicx} 
\usepackage{highlight}
\usepackage{caption}
\usepackage{tcolorbox}
\usepackage{fancyvrb}
\usepackage{fvextra}
\usepackage{wrapfig}
\usepackage{bbm}
\usepackage[flushleft]{threeparttable}
\usepackage{enumitem}
\usepackage{verbatim}
\usepackage{scalerel} 

\usepackage{subfig}

\usepackage{hyperref}
\hypersetup{
 colorlinks=true,
 linkcolor=cyan,
 filecolor=blue,      
 urlcolor=red,
 citecolor=cyan,
}

\usepackage{listings}
\usepackage{xcolor}

\definecolor{codegreen}{rgb}{0,0.6,0}
\definecolor{codegray}{rgb}{0.5,0.5,0.5}
\definecolor{codepurple}{rgb}{0.58,0,0.82}
\definecolor{backcolour}{rgb}{0.95,0.95,0.92}

\lstdefinestyle{mystyle}{
    backgroundcolor=\color{backcolour},   
    commentstyle=\color{codegreen},
    keywordstyle=\color{magenta},
    numberstyle=\tiny\color{codegray},
    stringstyle=\color{codepurple},
    basicstyle=\ttfamily\footnotesize,
    breakatwhitespace=false,         
    breaklines=true,                 
    captionpos=b,                    
    keepspaces=true,                 
    numbers=left,                    
    numbersep=5pt,                  
    showspaces=false,                
    showstringspaces=false,
    showtabs=false,                  
    tabsize=2
}

\begin{document}

\title{
Towards End-to-End Embodied Decision Making via Multi-modal Large Language Model: Explorations with GPT4-Vision and Beyond}

\preprintcopy

\author{Liang Chen$^1$, Yichi Zhang$^1$, Shuhuai Ren$^1$, Haozhe Zhao$^1$, Zefan Cai$^1$, \textbf{Yuchi Wang}$^1$\\
\textbf{Peiyi Wang}$^1$, \textbf{Tianyu Liu}$^2$, \textbf{Baobao Chang}$^1$\thanks{Corresponding author.}\\
$^1$ Peking University, $^2$ Tencent Cloud AI \\
 \texttt{\{leo.liang.chen, yczhang, shuhuai\_ren\}@stu.pku.edu.cn}  \\
 \texttt{\{hanszhao, zefan, wangyuchi, wangpeiyi\}@stu.pku.edu.cn}  \\
 \texttt{\{tianyu0421, chbb\}@pku.edu.cn}  \\
 \textbf{\url{https://github.com/pkunlp-icler/PCA-EVAL}}
}

\maketitle
\begin{abstract}

In this study, we explore the potential of Multimodal Large Language Models (MLLMs) in improving embodied decision-making processes for agents. While Large Language Models (LLMs) have been widely used due to their advanced reasoning skills and vast world knowledge, MLLMs like GPT4-Vision offer enhanced visual understanding and reasoning capabilities. We investigate whether state-of-the-art MLLMs can handle embodied decision-making in an end-to-end manner and whether collaborations between LLMs and MLLMs can enhance decision-making. To address these questions, we introduce a new benchmark called \textbf{PCA-EVAL}, which evaluates embodied decision-making from the perspectives of \textbf{P}erception, \textbf{C}ognition, and \textbf{A}ction. Additionally, we propose \textbf{HOLMES}, a multi-agent cooperation framework that allows LLMs to leverage MLLMs and APIs to gather multimodal information for informed decision-making. We compare end-to-end embodied decision-making and HOLMES on our benchmark and find that the GPT4-Vision model demonstrates strong end-to-end embodied decision-making abilities, outperforming GPT4-HOLMES in terms of average decision accuracy (+3\%). However, this performance is exclusive to the latest GPT4-Vision model, surpassing the open-source state-of-the-art MLLM by 26\%. Our results indicate that powerful MLLMs like GPT4-Vision hold promise for decision-making in embodied agents, offering new avenues for MLLM research. 

\end{abstract}

\section{Introduction}
The capacity to make well-informed decisions is essential for the survival and success of living organisms in their respective environments. Similarly, a major goal in embodied artificial intelligence is to develop agents, like robots, with sophisticated decision-making abilities. This could enable artificial agents to intelligently interact with their surroundings and efficiently accomplish a variety of real-world tasks such as autonomous driving~\citep{hu2023_uniad,lingo}, domestic assistance~\citep{ai2thor,ALFRED20,huang2022inner}, and game playing~\citep{fan2022minedojo,wang2023voyager,zhu2023ghost}. Recently, there has been a notable increase in leveraging exceptional reasoning capabilities and world knowledge of Large Language Models (LLMs) to enhance decision making in agents. However, LLMs are primarily designed to process textual context, creating a modality gap~\citep{Liang2022MindTG, ren-etal-2023-delving} for the LLM-powered agent when dealing with multimodal observations in real-world scenarios. 

To bridge this modality gap, a common approach is converting multimodal observations into text using various APIs~\citep{Wu2023VisualCT, Yang2023MMREACTPC}. However, this conversion can result in information loss during the transition from multimodal to unimodal text. 
At the same time, recent advances in Multimodal Large Language Models (MLLMs), particularly Visual Large Language Models (VLLMs) like GPT4-Vision~\citep{GPT-4V}, have showcased impressive general-purpose visual understanding and reasoning abilities~\citep{zhu2023minigpt,dai2023instructblip,liu2023llava,li2023m3it,zhao2023mmicl}. These VLLMs can directly perceive the visual information rather than relying on textual intermediaries, potentially enabling more sophisticated reasoning and decision making for embodied agents operating in complex real-world environments. 
Considering these developments, two research questions naturally arise: 
\textbf{(1)} Can current state-of-the-art VLLMs perform various embodied decision making tasks in an end-to-end manner? What are the current strengths and limitations when compared to LLM-powered agents? \textbf{(2)} Can LLMs and VLLMs collaborate to enhance embodied decision-making capabilities? 

\begin{figure}[t]
    \centering
    \includegraphics[width=0.6\textwidth]{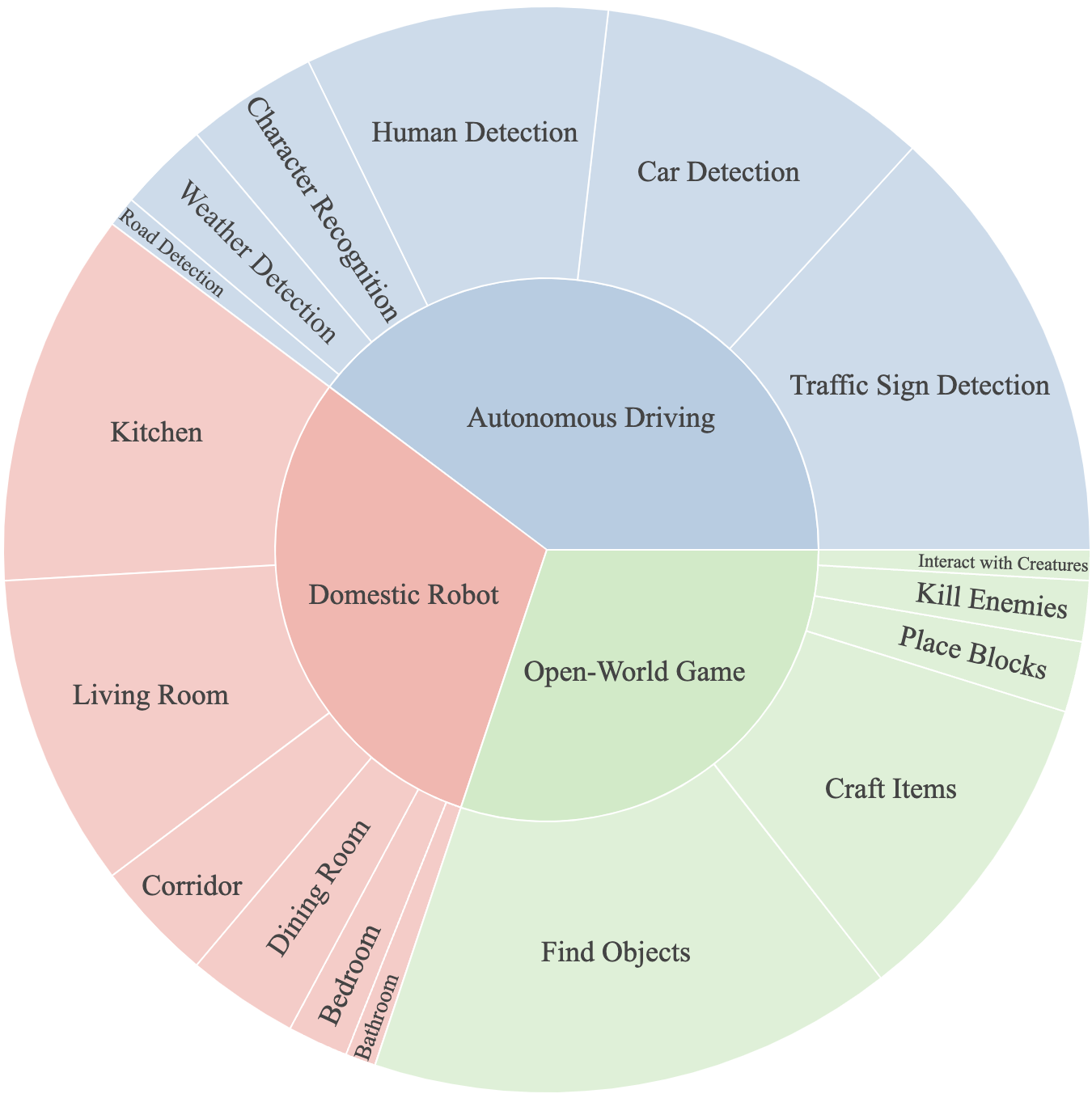}
    \caption{Domain and required ability distribution of PCA-EVAL.
    }
    \label{fig:sun-figure}
\end{figure}

However, addressing these questions is challenging due to the absence of an existing evaluation benchmark that satisfies the following criteria: \textbf{(1)} supporting end-to-end embodied decision making by providing agents with direct multimodal observations; \textbf{(2)} enabling multi-dimensional evaluation of the decision-making process, encompassing perception, reasoning, and action perspectives, rather than relying solely on final rewards or success rate; and \textbf{(3)} covering diverse domains, drawing from different areas of embodied AI. 
The development of more comprehensive benchmarks that meet these desiderata could substantially advance research on decision making in embodied systems.

In this paper, we propose a new benchmark, \textbf{PCA-EVAL}, for evaluating the embodied decision-making ability of agents from three perspectives, i.e.,  \textbf{P}erception, \textbf{C}ognition, and \textbf{A}ction. Our benchmark covers three domains as illustrated in Figure~\ref{fig:sun-figure}: autonomous driving, domestic assistance, and game-playing. The corresponding data are collected from real-world transportation scenes~\citep{Zhe_2016_CVPR_traffic_sign}, domestic housekeeper environment based on ALFRED~\citep{ALFRED20}, and Open-world environment Minedojo~\citep{fan2022minedojo} based on the famous game Minecraft. This diverse set of domains allows for a comprehensive assessment of embodied decision-making capabilities across various contexts. Distinct from the MDP-based evaluation that solely focuses on maximizing cumulative rewards, we divide the sequential decision making process into multiple one-step decision problems based on a task-specific topology graph. Each instance in the benchmark consists of a 6-element tuple: $<\textit{image},~\textit{question},~\textit{action candidates},~\textit{answer},~\textit{reason},~\textit{key concept}>$. Adopting this approach offers two major advantages: \textbf{(1)} It enables a more comprehensive evaluation of the decision-making process, with each decision step being assessed in terms of perception, cognition, and action. \textbf{(2)} The evaluation can be conducted outside complex simulation environments, simplifying the process of evaluating different agents and their performance.

With the proposed benchmark, we conduct two series of evaluation: (1) We examine multiple state-of-the-art VLLMs, like InstructBLIP~\citep{dai2023instructblip},  MMICL~\citep{zhao2023mmicl}, QwenVL-Chat~\citep{bai2023qwenvl} and the latest GPT4-Vision~\citep{GPT-4V}, in an end-to-end decision making context.  (2) We introduce \textbf{HOLMES},\footnote{The system is aptly named after the renowned detective, Sherlock Holmes.} a multi-agent cooperation framework. In this framework, we provide large language models, such as ChatGPT~\citep{chatgpt}, GPT4~\citep{openai2023gpt4}, and Vicuna~\citep{vicuna2023}, with descriptions of vision models like image captioning, object detection, Optical Character Recognition (OCR), and traffic sign detection models. 
Additionally, we supply descriptions of valid APIs within the simulated environment. The large language model subsequently initiates a search for clues pertaining to the question by engaging in a multi-turn conversation. This process involves alternating between invoking models or APIs to find clues and analyzing the discovered clues to facilitate informed decision making.

From our experimental results, we discerned that within the end-to-end framework, GPT4-Vision significantly outshines the contemporary state-of-the-art vision-language model, MMICL, boasting an average action accuracy improvement of 26\%. Notably, GPT4-Vision can furnish a detailed rationale behind its embodied decision-making process, a feature absent in present open-source VLLMs. When assessing HOLMES models, GPT4 consistently emerges superior across all three domains. Drawing a comparison between GPT4-Vision and HOLMES, we observed that GPT4-Vision surpasses GPT4-HOLMES with multiple expert visual APIs in terms of cognition and action scores. This underscores its broad adaptability across a spectrum of visual tasks and its good fusion of visual understanding, world knowledge, and embodied decision making.

In summary, we introduce three key contributions in this study:
\begin{enumerate}
    \item We propose PCA-EVAL, a novel evaluation benchmark for multi-domain embodied decision making that evaluates performance in perception, cognition, and action.
    \item We present HOLMES, a multi-agent cooperation framework designed to tackle various embodied decision-making tasks that include multimodal observations. It mimics the process of playing a detective game in which the LLM uncovers clues by utilizing various multimodal models or APIs supplied by the environment. 
    \item We conducted a systematic comparison of two embodied decision-making methods: end2end and HOLMES, across various models. Our findings suggest that when utilizing MLLM with the end2end method, it not only achieves decision accuracy better than the top-performing model (GPT-4) in HOLMES but also secures a superior cognition score. However, this level of performance is exclusive to the latest GPT4-Vision model, which significantly outpaces the open-source state-of-the-art VLLMs.
\end{enumerate}

We believe that powerful MLLMs like GPT4-Vision pave a new and promising way toward decision making in embodied agents using LLMs. It enables decisions across diverse domains to be made and justified seamlessly in an end-to-end manner.  PCA-EVAL serves as an effective metric for evaluating the embodied decision-making capabilities of both end-to-end and HOLMES-based models.
\section{Related Work}

\paragraph{Embodied Decision Making.}
Research on embodied decision-making is an emerging trend for artificial intelligent agents to interact with their surroundings and accomplish numerous tasks. This necessitates proficiency in vision perception, world knowledge, and commonsense reasoning, areas where a large language model can provide some level of expertise. 
We group prior work on embodied decision-making with LLM into two main trends. 
The first trend is to transform multimodal information, including object and scenery identification, the current states of AI agents, and the feedback from the environments, to texts. Text-based LLMs can then reason over the textual clues to determine the next action towards completing a designated task~\citep{huang2022language,li2022pre,huang2022inner,chen2023asking}. This line of research divides the entire decision-making process into two phases: (1) information seeking, usually involving VLLMs to verbalize the current status of AI agents in the vision-based environment with natural language; (2) reasoning and planning with text-based LLMs to decide what the AI agent should do in the next step with textual clues. 
The other line of research uses multimodal LLMs directly for end-to-end decision making, such as PALM-E~\citep{driess2023palm}.
The end-to-end decision making poses greater challenges to multimodal LLMs as it requires the combination of different functionalities including perception, cognition, and action, whereas decision making without explicit multiple steps mitigates the error propagation between information seeking and reasoning.








\paragraph{LLM-Powered Agents.}

Large language models pre-trained on large-scale multimodal (including text, image, video, etc.) corpus demonstrate impressive emergent abilities and immense popularity~\citep{brown2020language,wei2022emergent}, and have seen tremendous success across various domains covering various natural language processing and computer vision tasks~\citep{radford2019language,chowdhery2022palm,touvron2023llama1,alayrac2022flamingo,zhu2023minigpt,li2023blip}. 
Consequently, using LLMs to empower the AI agents~\citep{xi2023rise,liu2023training,park2023generative,Wang2023DescribeEP,yuan2023plan4mc} becomes more and more promising. 
Specifically, we can employ LLMs to enhance the decision making ability of the agents~\citep{nakano2022webgpt,yao2022react,li2023apibank,song2023restgpt}, expanding their perception and action space through strategies like tool utilization~\citep{schick2023toolformer,qin2023tool,lu2023chameleon}.
Although LLM-based agents demonstrate reasoning and planning abilities through techniques like Chain of Thought or problem decomposition~\citep{wei2023chainofthought,yao2023retroformer,kojima2022large}, they inherently lack visual perception, and are limited to the discrete textual content. Therefore, integrating visual information or other modalities can offer agents a broader context and a more precise understanding~\citep{driess2023palme}, enhancing their environmental perception. However, no evaluation protocol or benchmark is currently available to evaluate decision making within the multimodal context.


\section{PCA-EVAL}

In this section, we propose to evaluate the decision-making ability of embodied agents from three perspectives: perception, cognition, and action. Accordingly, we present a novel benchmark named PCA-EVAL. 
Our PCA-EVAL benchmark consists of 300 multimodal multiple-choice questions with diverse embodied topics and annotations of their answers with corresponding explanations. 

As shown in Figure~\ref{fig:example-pcaeval}, each instance in the benchmark consists of a 6-element tuple: $<$\textit{image}, \textit{question}, \textit{action candidates}, \textit{answer}, \textit{reason}, \textit{key concept}$>$. The image is collected from various embodied environments, like transportation scenes, housekeeper environments, and game worlds in Minecraft. Questions, action candidates, and answers are derived from real tasks within the corresponding environment. The reasoning explains why the answer is the best choice for the current image, while the key concept highlights the most question-related aspect in the image.

\begin{wrapfigure}{r}{0.4\textwidth}
  \vspace{-5mm}
  \begin{center}
    \includegraphics[width=0.4\textwidth]{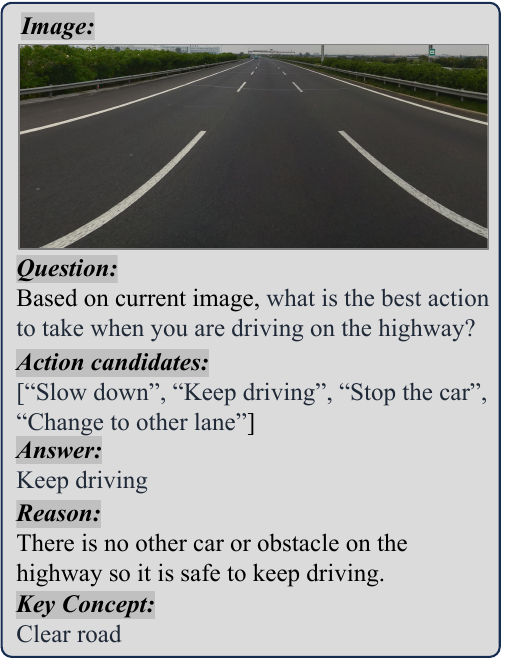}
  \end{center}
  \vspace{-0.4cm}
  \caption{An instance of PCA-EVAL.}
  \label{fig:example-pcaeval}
  \vspace{-6mm}
\end{wrapfigure}

Unlike traditional visual question-answering datasets that emphasize visual perception (e.g., VQA~\citep{balanced_vqa_v2}), visual reasoning (e.g., NLVR~\citep{Suhr2017NLVR}), or world knowledge (e.g., OKVQA~\citep{marino2019okvqa}), the most distinctive characteristic of PCA-EVAL is its grounding in embodied actions. Compared to embodied simulation environments like ALFRED~\citep{ALFRED20} and Minedojo~\citep{fan2022minedojo}, PCA-EVAL proves to be more effective in evaluating various LLM-based agents. This is primarily due to PCA-EVAL's provision of high-level actions that can be readily implemented or programmed using the low-level actions in the corresponding domains. The high-level actions are more comprehensible for LLMs than the direct low-level actions like robotic movements in the simulation environments because (1) the high-level actions are in the form of natural languages, making it easier for LLMs to understand the meaning and connect with world knowledge. (2) LLMs are not grounded with low-level actions during the pretraining or finetuning stage, making it hard for LLMs to understand the consequences of executing an action.

To answer a question in PCA-EVAL, the agent must possess the following abilities: (1) Perception: accurately identify the concept related to the question within the image; (2) Cognition: engage in reasoning based on image perception and worldly knowledge; (3) Action: comprehend the potential actions, selecting the one that best aligns with the outcome of the reasoning process. A deficiency in any of these abilities would inevitably result in an incorrect answer, posing a significant challenge to the more complex capabilities of embodied agents. Although challenging, all the aforementioned abilities are essential for the decision-making process in embodied environments.

\subsection{Evaluation Metrics}

For each instance, we instruct the agent to deliver an answer triplet comprising an image description $d$, a reasoning process $r$, and a final action $a$, represented as $<d,r,a>$. By comparing the model prediction with the ground truth answer, we can obtain a fine-grained diagnosis of the decision making process. 


\paragraph{Perception Score.} The Perception Score (P-Score) measures the model's ability to accurately perceive and interpret the observation. It is computed based on whether the agent's output image description $d$ includes the key concept of the instance. If the agent accurately describes the question-related key concept in the image, the P-score is assigned a value of 1; otherwise, it is assigned a value of 0. For the instance in Figure~\ref{fig:example-pcaeval}, the agent should output ``clear road'' or ``no car visible'' or other semantically equivalent concepts in its description of the image to get the perception score.


\paragraph{Cognition Score.} The Cognition Score (C-Score) assesses the model's ability to reason, comprehend, and make informed decisions based on the perceived input data and world knowledge. The score is 1 if the reasoning process is correct, otherwise the score is 0. For the instance in Figure~\ref{fig:example-pcaeval}, the agent should link the ``clear road'' to the action ``keep driving'' based on transportation commonsense to get the score.

\paragraph{Action Score.} The Action Score (A-Score) measures the model's ability to generate appropriate and effective responses or actions based on the perceived input data and the cognitive understanding of the context. The score is assigned a value of 1 if the agent selects the correct action; otherwise, the score is set to 0. 

The final Perception, Cognition, and Action scores of the agents are obtained by averaging the scores across all instances and domains in our PCA-EVAL dataset. 


\subsection{Automatic Evaluation}

Recent advancements have seen researchers harnessing powerful LLMs for the evaluation of output of language models. Studies have revealed that the outcomes from LLMs could exhibit remarkable alignment with human judgments \cite{zheng2023judging, wang2023large, wang2023making}. In our investigation, we employed GPT-4 to automatically evaluate perception, cognition, and action scores based on the model's outputs. Our findings underscore a significant agreement between GPT-4 annotations and human annotator results. This is substantiated by Pearson correlation coefficients of 0.8, 0.9, and 0.95 for perception, cognition, and action evaluations, respectively. To facilitate ongoing and future research endeavors, we share our automatic evaluation script\footnote{\url{https://github.com/pkunlp-icler/PCA-EVAL/blob/main/pca-eval/evaluation/pca_auto_scoring.py}} for seamless adoption, which could also be improved in the future. For a detailed description of our evaluation methodology, kindly refer to Appendix \ref{app:ae}

\subsection{Dataset Overview}

The PCA-EVAL benchmark currently comprises three domains, with a total of 300 instances, including 100 instances per domain.
In our preliminary study, we find that the annotation process requires proactive thinking of the questions, actions, and corresponding answers, which makes quality control difficult. In order to ensure the quality of PCA-Eval, every single test case has been verified by at least three authors of this paper. Although challenging, we would keep scaling this benchmark in order to advocate further attention to end-to-end decision-making. We introduce the three domains encompassed by our dataset as follows:

\paragraph{Autonomous Driving.} In the autonomous driving domain, instances are derived from real-world transportation scenes, which requires the agent to have particular abilities such as traffic sign recognition, obstacle detection, and decision-making at intersections. The dataset aims to evaluate an agent's ability to perceive and interpret visual information while making safe and efficient driving decisions. The images are collected from TT100K~\citep{Zhe_2016_CVPR_traffic_sign} dataset and annotators are instructed to propose an image-conditioned question that is grounded with real actions of vehicles. 

\paragraph{Domestic Robot.} The domestic assistance domain features instances from the ALFRED~\citep{ALFRED20,ai2thor} environment, which simulates a housekeeper robot performing tasks within a household setting. These tasks may include object manipulation, navigation, and interaction with various appliances. The environment assesses an agent's ability to understand and execute complex instructions while navigating and interacting with a dynamic environment. Annotators are asked to select one image from the randomly generated scenes in the environment, propose a question related to the items on the scene, and annotate the full information of the instance. 

\begin{wrapfigure}{r}{0.4\textwidth}
\vspace{-6mm}
  \begin{center}
    \includegraphics[width=0.4\textwidth]{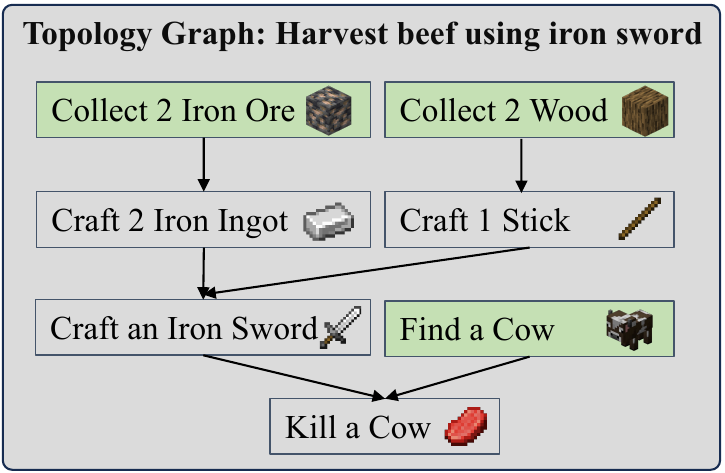}
  \end{center}
  \vspace{-0.4cm}
  \caption{Illustration of task topology graph. Events in green represent the leaf nodes of the graph.}
  \label{fig:topology}
  \vspace{-5mm}
\end{wrapfigure}

\paragraph{Open-World Game.} In the open-world game domain, instances are sourced from the Minecraft environment, where agents are tasked with exploring, crafting, and surviving in a procedurally generated world. This dataset evaluates an agent's ability to reason and plan actions within a complex, open-ended environment, which often requires long-term strategizing and adaptability. Annotators receive predefined tasks from MineDojo~\citep{fan2022minedojo} as a reference during the task generation phase. For each task, we instruct the annotator to sketch a task topology graph, exemplified in Figure~\ref{fig:topology}. The task should be completed in accordance with the topological order of the graph, where the event located in the leaf nodes should be finished first. Each node in the task topology graph can be viewed as a step in the sequential decision. We list the in-domain task distribution and examples for each domain in Appendix~\ref{app:example_pca_eval}.

\subsection{Annotation Pipelines}

The annotation process consists of two stages: (1) Dataset Annotation, and (2) Dataset Refinement. During the initial stage, three annotators are assigned to each domain, adhering strictly to the respective annotation guidelines. They first pinpoint the source images from each domain that are informative and meaningful so that they can write questions for each image. 
The annotators have the responsibility to ensure every question has only one correct answer and accurate rationales. In the subsequent stage, annotators are instructed to scrutinize the output actions and rationales presented by ChatGPT and check the annotations. This process aims to address the challenge of multiple correct answers, as ChatGPT can furnish comprehensive explanations for its actions. These explanations assist annotators in assessing the acceptability of ChatGPT's response, particularly when it deviates from the established ground truth answer. This enables annotators to refine annotations to ensure the presence of a single correct answer.


\section{Methods}

\subsection{End2End Decision Making via VLLMs}

\begin{figure}[t]
    \centering
    \includegraphics[width=\textwidth]{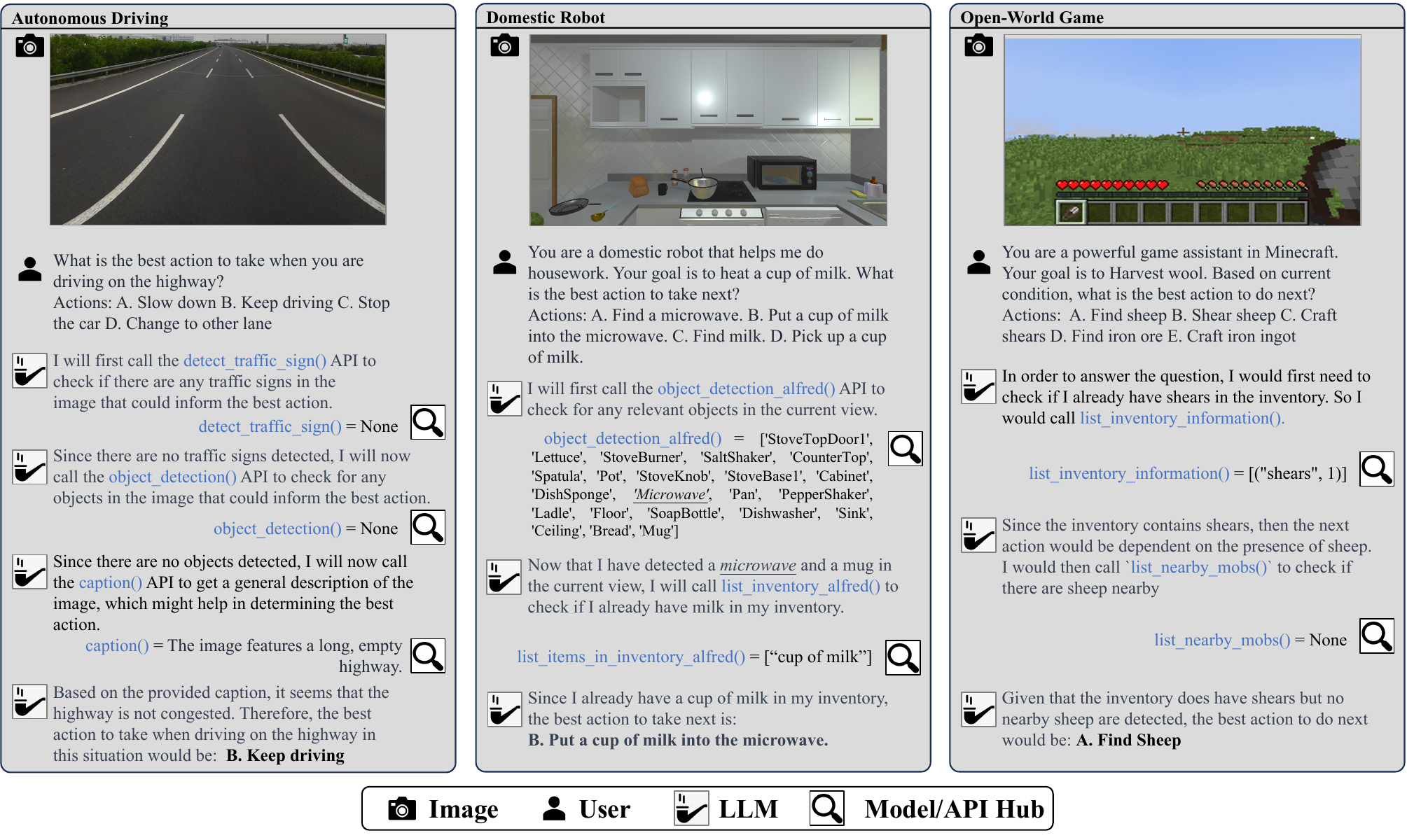}
    \caption{Three examples of HOLMES solving questions from different domains of PCA-EVAL.
    }
    \label{fig:illustrative-example}
\end{figure}

In this subsection, we detail the evaluation process for assessing state-of-the-art VLLMs, e.g., InstructBLIP,  MMICL, and GPT4-Vision, on end-to-end embodied decision-making using the proposed PCA-EVAL benchmark. End2End embodied decision making is straightforward since we can directly feed the visual observation and the textual question to the multi-modal agent. As illustrated in Figure~\ref{fig:example-pcaeval}, the agent is prompted to output the image description and reasoning process before giving the final action. 

\begin{wrapfigure}{r}{0.4\textwidth}
\vspace{-15mm}
  \begin{center}
    \includegraphics[width=0.4\textwidth]{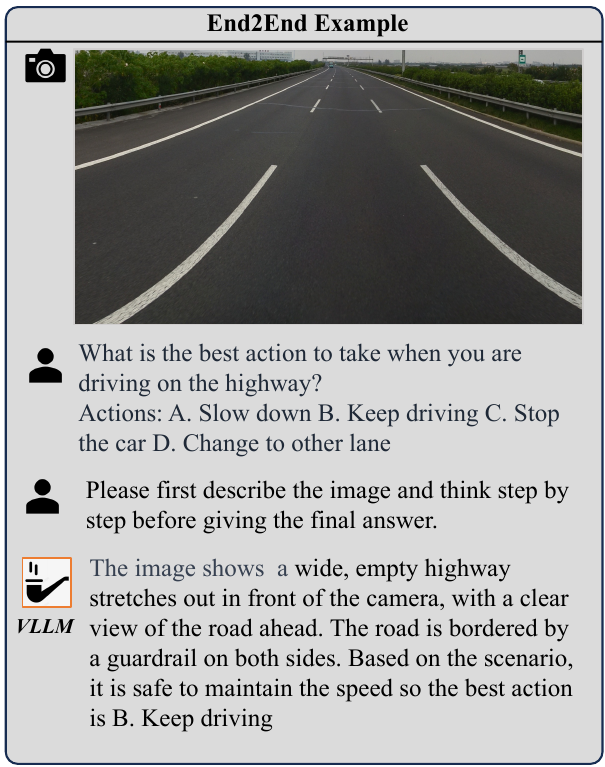}
  \end{center}
  \vspace{-0.4cm}
  \caption{An example of end-to-end decision making.}
  \label{fig:example-pcaeval}
  \vspace{-6mm}
\end{wrapfigure}

\subsection{HOLMES: Multi-Agent Cooperation}

Different from End2End embodied decision making, within HOLMES, we prompt large language models like ChatGPT-3.5~\citep{chatgpt}, GPT4~\citep{openai2023gpt4} to call different visual models or APIs to gather information about the environment.

We provide these models with descriptions of the input and output for different visual models such as the image caption model based on InstructBLIP, the object detection model based on POMP~\citep{ren2023pomp}, and the traffic sign detection model based on YOLO~\citep{YOLO}. Additionally, we supply descriptions of valid APIs within the simulated environment, such as \textit{list\_nearby\_mobs\_in\_minecraft()} to tell what creatures can current player see and \textit{list\_items\_at\_hand\_in\_alfred()} to tell what item the robot is holding in hand. Full API description files for each domain are shown in Appendix~\ref{app:api_des}.

These integrations enable the large language model to initiate a search for clues pertaining to a given question through a multi-turn conversation. As shown in Figure~\ref{fig:illustrative-example}, the process involves alternating between invoking models or APIs to gather relevant information and analyzing the discovered clues to facilitate informed decision making. The HOLMES framework is designed to enhance cooperation and coordination among multiple agents in dynamic and complex environments. 

In HOLMES, there are four key components as depicted in Figure~\ref{fig:illustrative-example}: the image, the user, the LLM, and the Model/API Hub. Initially, the user poses a question about the optimal action to take based on the environment shown in the image, providing potential action choices. As the LLM cannot directly view the image, it's briefed with descriptions of available visual models and APIs supplied by the simulation environment. It's then tasked with gathering relevant data via these models and APIs to determine the appropriate action. When the LLM responds, the system checks if it has invoked a legitimate model or API, subsequently relaying the results from the invoked API. This feedback is logged into the dialogue history, allowing the LLM to analyze and form subsequent responses. Once equipped with sufficient information, the LLM proposes the final action, accompanied by its underlying rationale. HOLMES emulates the detective game process, where one alternates between searching for clues using various tools and analyzing them before arriving at a conclusion.

\section{Experiments}

\subsection{Configurations}
\paragraph{End2End.}

Under this setting same image and prompts are provided to different VLLMs. Additionally, the non-visual information ``items in hand''  and ``items in inventory'' for domestic and game domains are directly given to the models in the prompt since these information is hard to perceive from the image and is easy to obtain from the simulation environments. We would also make the prompts we use open-source for fair and convenient evaluation.

We compare four different models, InstructBLIP-Vicuna-13B\footnote{\url{https://github.com/salesforce/LAVIS/tree/main/projects/instructblip}}, MMICL-FLANT5XXL\footnote{\url{https://huggingface.co/BleachNick/MMICL-Instructblip-T5-xxl}}, QwenVL-Chat\footnote{\url{https://huggingface.co/Qwen/Qwen-VL-Chat}} and GPT4-Vision\footnote{\url{https://chat.openai.com}}. We apply default inference configurations for the corresponding models. 

\paragraph{HOLMES.}

In HOLMES framework, the LLM is required to continuously invoke various APIs and retrieve their return information. To streamline the evaluation process, we initially execute all APIs for every instance in PCA-EVAL, storing the result for each instance. This approach allows us to directly access the specific result of a given API without the need to run the model each time an evaluation is conducted. We would also make the API results open-source together with the benchmark. The description and implementation details of the APIs are listed in Appendix~\ref{app:api_des}.

We compare three LLMs: Vicuna\footnote{\url{https://huggingface.co/lmsys}}, ChatGPT-3.5-Turbo and GPT4\footnote{\url{https://platform.openai.com}}. However we found Vicuna models lack the capability to call various APIs for information gathering, thus we have only reported the results for ChatGPT and GPT4. We anticipate supplementing these results as soon as open-source models become available, which can understand API descriptions and correspondingly call different APIs.

\subsection{Evaluation}

PCA-Eval assesses embodied decision-making through three distinct lenses: perception, cognition, and action. The scores we reported in Table~\ref{tab:main} rely on the consensus score from three human evaluators. We compute the average kappa correlation coefficient for these evaluators, resulting in 0.91 for the Perception Score and 0.88 for the Cognition Score. These figures 
indicate a good consistency in the evaluation process.

\subsection{Main Results}

\begin{table}[t]
    \centering
    \resizebox{\textwidth}{!}{
    \begin{tabular}{c|c|ccc|ccc|ccc|ccc}
    \toprule
        \multirow{2}{*}{Method} & \multirow{2}{*}{Model}  & \multicolumn{3}{c|}{Traffic} & \multicolumn{3}{c|}{Domestic}  & \multicolumn{3}{c|}{Game} & \multicolumn{3}{c}{Average} \\ 
        ~ & ~  & P & C & A & P & C & A & P & C & A &P & C & A \\
        \midrule
        \multirow{4}{*}{End2End} 



        &InstructBLIP\color{red}{$^\dagger$}  & - & - & 0.42 & - & - & 0.41 & - & - & 0.24 & - & - &0.36  \\ 
        ~ & MMICL\color{red}{$^\dagger$}  & - & - & 0.63 & - & - & 0.51 & - & - & 0.29 & - & - & 0.48 \\ 

        ~ & QwenVL-Chat\color{red}{$^\dagger$}  & - & - & 0.59 & - & - & 0.55 & - & - & 0.24 & - & - & 0.46  \\

        ~ & GPT-4V$^\ddagger$  & 0.75 & 0.73 & 0.78 & 0.81 & \textbf{0.69} & \textbf{0.67} & \textbf{0.95} & \textbf{0.79} & \textbf{0.77} &0.84 & \textbf{0.74} & \textbf{0.74} \\ 
        \midrule
        
        \multirow{2}{*}{HOLMES} & ChatGPT$^\ddagger$  & 0.75 & 0.68 & 0.66 & \textbf{0.88} & 0.52 & 0.50 & 0.78 & 0.40 & 0.36 & 0.80 & 0.53 & 0.51 \\ 
        ~ & GPT4$^\ddagger$  & \textbf{0.87} & \textbf{0.82} & \textbf{0.82} & 0.85 & 0.61 & 0.56 & 0.91 & 0.77 & 0.74 & \textbf{0.88} & 0.73 & 0.71 \\ 
        \bottomrule
    \end{tabular}}
    \caption{Main results on PCA-EVAL. Models with \color{red}{$\dagger$} \color{black} are fully open-source. Models with $\ddagger$ only provide API to access. P, C, and A represent Perception, Cognition, and Action Scores, respectively. For the open-source models in End2End setting, we find it hard to prompt them to output correct cross-modal reasoning information, so their Perception and Cognition scores are not reported. }
    \label{tab:main}
\end{table}



We evaluate various methods and models on the PCA-EVAL benchmark, as shown in Table~\ref{tab:main}.

In the upper block concerning End2End-VLLMs, the recently unveiled closed-source model, GPT-4V, outperforms existing open-source models by achieving the highest scores of 0.84, 0.74, and 0.74 in the perception, cognition, and action dimensions respectively. This performance represents a 26\% action score improvement over its open-source counterpart, MMICL. The impressive performance of GPT-4V is primarily attributed to its exceptional ability to perceive visual information across different domains, particularly in the challenging game domain.

We also assessed the performance of embodied decision making using our HOLMES system.

As shown in the bottom block of the table, the HOLMES system, based on GPT4, achieves an Action Score of 0.71, matching the performance of GPT-4V (0.74). This suggests that the HOLMES system is proficient in understanding the task goal, breaking down the larger goal into multiple smaller steps, and accurately invoking the relevant APIs to accomplish each step.

Specifically, the GPT4-HOLMES system can identify key concepts in an image through the results returned by APIs such as \textit{list\_nearby\_mobs\_in\_minecraft()}. As a result, the system achieves an average Perception Score of 0.88, surpassing GPT-4V's 0.84. However, when compared to End2End methods, HOLMES relies on multi-step reasoning for the final decision. This approach can lead to the accumulation of reasoning errors, resulting in a lower Cognition Score in both Domestic and Game domains.


\section{Discussion}

\subsection{Comparison Between End2End and HOLMES}

\begin{figure}[t!]
    \centering
    \subfloat[A Comparison between GPT4-V and GPT4-HOLMES]{\includegraphics[height=0.55\textwidth]{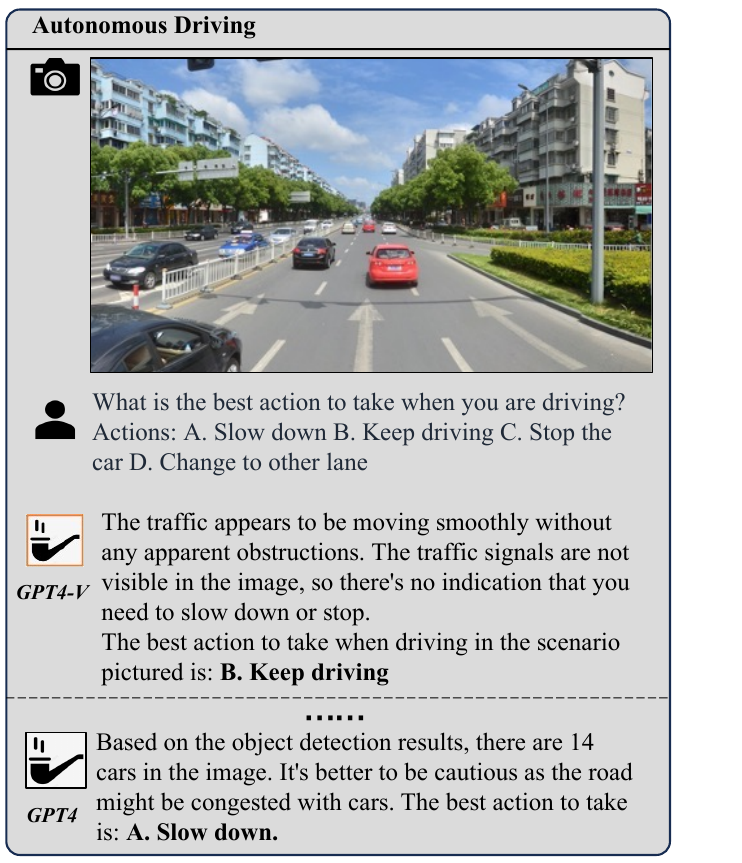}\label{fig:compare}}
\hfill
    \subfloat[An example of decision misalignment.]{\includegraphics[height=0.55\textwidth]{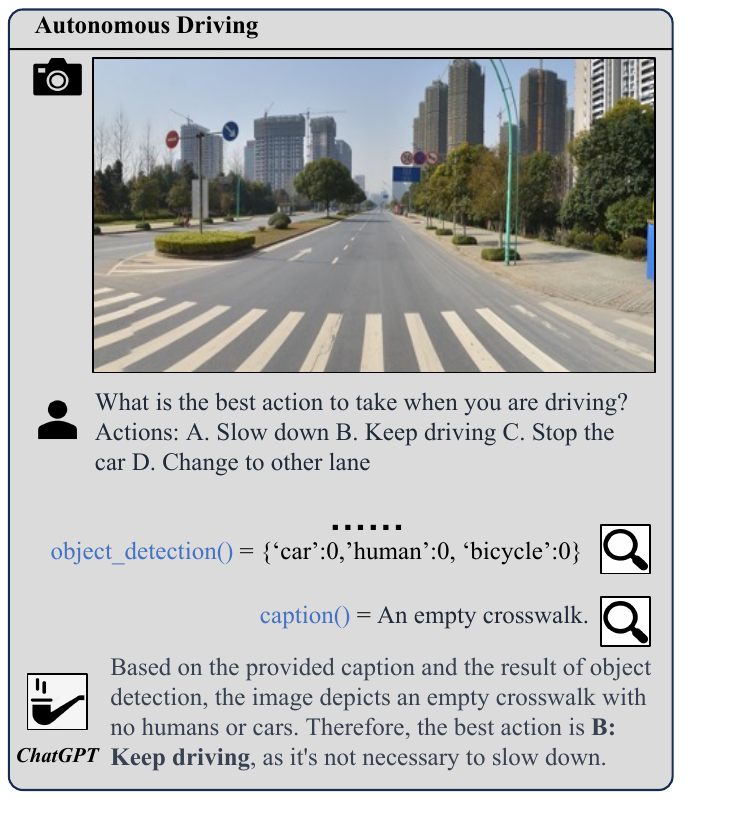}\label{fig:alignment}}
    \caption{Case studies.}

\end{figure}

We conduct an analysis and comparison of the outputs generated by the End2End method with GPT4-Vision, as well as the HOLMES method with GPT4. Our findings indicate that the End2End method effectively mitigates information loss during the modality conversion process. As illustrated in Figure~\ref{fig:compare}, an image depicts a road with several nearby cars. GPT4-Vision is capable of discerning that these cars are situated in a safe space, thereby suggesting that the driver can continue driving.

Conversely, GPT4, while aware of the number of cars, lacks information about their spatial relation, leading it to recommend slowing down. This suggests that the End2End method is superior in perceiving certain visual features that are not captured by the APIs. Conversely, some specialized APIs, such as traffic sign detection, outperform GPT4-Vision in tasks like traffic sign detection, as they are specifically trained for this task. This could enable the HOLMES method to gather more accurate information than the End2End model.

\subsection{Alignment between Agent Decisions and Human Values}


We have observed instances where the decisions made by the agent contradict human values. For instance, consider the scenario depicted in Figure~\ref{fig:alignment}. The image illustrates a crosswalk devoid of pedestrians. The appropriate response in this situation would be to slow down, as caution is paramount when approaching a crosswalk, regardless of the presence or absence of pedestrians. However, upon processing the information that the crosswalk is unoccupied, ChatGPT suggests that maintaining the current speed is the optimal action, arguing that the absence of pedestrians eliminates the need to slow down. The rationale provided by ChatGPT is logical, yet it does not align with human values. We believe it is crucial for embodied agents to make decisions that are in harmony with human values, rather than solely focusing on maximizing their advantage.

\subsection{Limitation and Future Work}

The current scope of PCA-EVAL is confined to merely three domains, with a cap of 100 instances per domain. One of our future work aims to broaden this scope to encompass more domains and embodied  environments where MLLMs could keep getting feedback. Furthermore, we plan to increase the number of instances for both the existing and newly introduced domains.


\section{Conclusion}

 
In this study, we present PCA-EVAL, a comprehensive evaluation benchmark for embodied decision-making that gauges performance in perception, cognition, and action, thereby offering an all-encompassing measure for various embodied agents. We conduct a systematic comparison between End2End embodied decision-making and HOLMES, a multi-agent cooperation framework developed by us. Our findings reveal that MLLM, when applied with the end2end method, surpasses the top-performing model in HOLMES, GPT-4, in terms of decision accuracy and cognition score. However, it is crucial to underscore that this superior performance is specific to the GPT4-Vision model, which significantly outperforms the open-source state-of-the-art VLLMs. These results and subsequent analysis underscore the necessity for ongoing exploration in embodied decision-making and the development of open-source MLLMs to ensure wider accessibility and progress in the field.


\newpage

\bibliography{custom,iclr2024_conference}

\begin{thebibliography}{58}
\providecommand{\natexlab}[1]{#1}
\providecommand{\url}[1]{\texttt{#1}}
\expandafter\ifx\csname urlstyle\endcsname\relax
  \providecommand{\doi}[1]{doi: #1}\else
  \providecommand{\doi}{doi: \begingroup \urlstyle{rm}\Url}\fi

\bibitem[Alayrac et~al.(2022)Alayrac, Donahue, Luc, Miech, Barr, Hasson, Lenc, Mensch, Millican, Reynolds, et~al.]{alayrac2022flamingo}
Jean-Baptiste Alayrac, Jeff Donahue, Pauline Luc, Antoine Miech, Iain Barr, Yana Hasson, Karel Lenc, Arthur Mensch, Katherine Millican, Malcolm Reynolds, et~al.
\newblock Flamingo: a visual language model for few-shot learning.
\newblock \emph{Advances in Neural Information Processing Systems}, 35:\penalty0 23716--23736, 2022.

\bibitem[Bai et~al.(2023)Bai, Bai, Yang, Wang, Tan, Wang, Lin, Zhou, and Zhou]{bai2023qwenvl}
Jinze Bai, Shuai Bai, Shusheng Yang, Shijie Wang, Sinan Tan, Peng Wang, Junyang Lin, Chang Zhou, and Jingren Zhou.
\newblock Qwen-vl: A versatile vision-language model for understanding, localization, text reading, and beyond, 2023.

\bibitem[Brown et~al.(2020)Brown, Mann, Ryder, Subbiah, Kaplan, Dhariwal, Neelakantan, Shyam, Sastry, Askell, Agarwal, Herbert-Voss, Krueger, Henighan, Child, Ramesh, Ziegler, Wu, Winter, Hesse, Chen, Sigler, Litwin, Gray, Chess, Clark, Berner, McCandlish, Radford, Sutskever, and Amodei]{brown2020language}
Tom~B. Brown, Benjamin Mann, Nick Ryder, Melanie Subbiah, Jared Kaplan, Prafulla Dhariwal, Arvind Neelakantan, Pranav Shyam, Girish Sastry, Amanda Askell, Sandhini Agarwal, Ariel Herbert-Voss, Gretchen Krueger, Tom Henighan, Rewon Child, Aditya Ramesh, Daniel~M. Ziegler, Jeffrey Wu, Clemens Winter, Christopher Hesse, Mark Chen, Eric Sigler, Mateusz Litwin, Scott Gray, Benjamin Chess, Jack Clark, Christopher Berner, Sam McCandlish, Alec Radford, Ilya Sutskever, and Dario Amodei.
\newblock Language models are few-shot learners, 2020.

\bibitem[Chen et~al.(2023)Chen, Zhang, Zhang, Zhao, and Chen]{chen2023asking}
Xiaoyu Chen, Shenao Zhang, Pushi Zhang, Li~Zhao, and Jianyu Chen.
\newblock Asking before action: Gather information in embodied decision making with language models.
\newblock \emph{arXiv preprint arXiv:2305.15695}, 2023.

\bibitem[Chiang et~al.(2023)Chiang, Li, Lin, Sheng, Wu, Zhang, Zheng, Zhuang, Zhuang, Gonzalez, Stoica, and Xing]{vicuna2023}
Wei-Lin Chiang, Zhuohan Li, Zi~Lin, Ying Sheng, Zhanghao Wu, Hao Zhang, Lianmin Zheng, Siyuan Zhuang, Yonghao Zhuang, Joseph~E. Gonzalez, Ion Stoica, and Eric~P. Xing.
\newblock Vicuna: An open-source chatbot impressing gpt-4 with 90\%* chatgpt quality, March 2023.
\newblock URL \url{https://lmsys.org/blog/2023-03-30-vicuna/}.

\bibitem[Chowdhery et~al.(2022)Chowdhery, Narang, Devlin, Bosma, Mishra, Roberts, Barham, Chung, Sutton, Gehrmann, et~al.]{chowdhery2022palm}
Aakanksha Chowdhery, Sharan Narang, Jacob Devlin, Maarten Bosma, Gaurav Mishra, Adam Roberts, Paul Barham, Hyung~Won Chung, Charles Sutton, Sebastian Gehrmann, et~al.
\newblock Palm: Scaling language modeling with pathways.
\newblock \emph{arXiv preprint arXiv:2204.02311}, 2022.

\bibitem[Dai et~al.(2023)Dai, Li, Li, Tiong, Zhao, Wang, Li, Fung, and Hoi]{dai2023instructblip}
Wenliang Dai, Junnan Li, Dongxu Li, Anthony Meng~Huat Tiong, Junqi Zhao, Weisheng Wang, Boyang Li, Pascale Fung, and Steven Hoi.
\newblock Instructblip: Towards general-purpose vision-language models with instruction tuning, 2023.

\bibitem[Driess et~al.(2023{\natexlab{a}})Driess, Xia, Sajjadi, Lynch, Chowdhery, Ichter, Wahid, Tompson, Vuong, Yu, Huang, Chebotar, Sermanet, Duckworth, Levine, Vanhoucke, Hausman, Toussaint, Greff, Zeng, Mordatch, and Florence]{driess2023palme}
Danny Driess, Fei Xia, Mehdi S.~M. Sajjadi, Corey Lynch, Aakanksha Chowdhery, Brian Ichter, Ayzaan Wahid, Jonathan Tompson, Quan Vuong, Tianhe Yu, Wenlong Huang, Yevgen Chebotar, Pierre Sermanet, Daniel Duckworth, Sergey Levine, Vincent Vanhoucke, Karol Hausman, Marc Toussaint, Klaus Greff, Andy Zeng, Igor Mordatch, and Pete Florence.
\newblock Palm-e: An embodied multimodal language model, 2023{\natexlab{a}}.

\bibitem[Driess et~al.(2023{\natexlab{b}})Driess, Xia, Sajjadi, Lynch, Chowdhery, Ichter, Wahid, Tompson, Vuong, Yu, et~al.]{driess2023palm}
Danny Driess, Fei Xia, Mehdi~SM Sajjadi, Corey Lynch, Aakanksha Chowdhery, Brian Ichter, Ayzaan Wahid, Jonathan Tompson, Quan Vuong, Tianhe Yu, et~al.
\newblock Palm-e: An embodied multimodal language model.
\newblock \emph{arXiv preprint arXiv:2303.03378}, 2023{\natexlab{b}}.

\bibitem[Fan et~al.(2022)Fan, Wang, Jiang, Mandlekar, Yang, Zhu, Tang, Huang, Zhu, and Anandkumar]{fan2022minedojo}
Linxi Fan, Guanzhi Wang, Yunfan Jiang, Ajay Mandlekar, Yuncong Yang, Haoyi Zhu, Andrew Tang, De-An Huang, Yuke Zhu, and Anima Anandkumar.
\newblock Minedojo: Building open-ended embodied agents with internet-scale knowledge.
\newblock In \emph{Thirty-sixth Conference on Neural Information Processing Systems Datasets and Benchmarks Track}, 2022.
\newblock URL \url{https://openreview.net/forum?id=rc8o_j8I8PX}.

\bibitem[Goyal et~al.(2017)Goyal, Khot, Summers{-}Stay, Batra, and Parikh]{balanced_vqa_v2}
Yash Goyal, Tejas Khot, Douglas Summers{-}Stay, Dhruv Batra, and Devi Parikh.
\newblock Making the {V} in {VQA} matter: Elevating the role of image understanding in visual question answering.
\newblock In \emph{2017 {IEEE} Conference on Computer Vision and Pattern Recognition, {CVPR} 2017, Honolulu, HI, USA, July 21-26, 2017}, pp.\  6325--6334, 2017.

\bibitem[Hu et~al.(2023)Hu, Yang, Chen, Li, Sima, Zhu, Chai, Du, Lin, Wang, Lu, Jia, Liu, Dai, Qiao, and Li]{hu2023_uniad}
Yihan Hu, Jiazhi Yang, Li~Chen, Keyu Li, Chonghao Sima, Xizhou Zhu, Siqi Chai, Senyao Du, Tianwei Lin, Wenhai Wang, Lewei Lu, Xiaosong Jia, Qiang Liu, Jifeng Dai, Yu~Qiao, and Hongyang Li.
\newblock Planning-oriented autonomous driving.
\newblock In \emph{Proceedings of the IEEE/CVF Conference on Computer Vision and Pattern Recognition}, 2023.

\bibitem[Huang et~al.(2022{\natexlab{a}})Huang, Abbeel, Pathak, and Mordatch]{huang2022language}
Wenlong Huang, Pieter Abbeel, Deepak Pathak, and Igor Mordatch.
\newblock Language models as zero-shot planners: Extracting actionable knowledge for embodied agents.
\newblock In \emph{International Conference on Machine Learning}, pp.\  9118--9147. PMLR, 2022{\natexlab{a}}.

\bibitem[Huang et~al.(2022{\natexlab{b}})Huang, Xia, Xiao, Chan, Liang, Florence, Zeng, Tompson, Mordatch, Chebotar, Sermanet, Brown, Jackson, Luu, Levine, Hausman, and Ichter]{huang2022inner}
Wenlong Huang, Fei Xia, Ted Xiao, Harris Chan, Jacky Liang, Pete Florence, Andy Zeng, Jonathan Tompson, Igor Mordatch, Yevgen Chebotar, Pierre Sermanet, Noah Brown, Tomas Jackson, Linda Luu, Sergey Levine, Karol Hausman, and Brian Ichter.
\newblock Inner monologue: Embodied reasoning through planning with language models.
\newblock In \emph{arXiv preprint arXiv:2207.05608}, 2022{\natexlab{b}}.

\bibitem[Kojima et~al.(2022)Kojima, Gu, Reid, Matsuo, and Iwasawa]{kojima2022large}
Takeshi Kojima, Shixiang~Shane Gu, Machel Reid, Yutaka Matsuo, and Yusuke Iwasawa.
\newblock Large language models are zero-shot reasoners.
\newblock \emph{Advances in neural information processing systems}, 35:\penalty0 22199--22213, 2022.

\bibitem[Kolve et~al.(2017)Kolve, Mottaghi, Han, VanderBilt, Weihs, Herrasti, Gordon, Zhu, Gupta, and Farhadi]{ai2thor}
Eric Kolve, Roozbeh Mottaghi, Winson Han, Eli VanderBilt, Luca Weihs, Alvaro Herrasti, Daniel Gordon, Yuke Zhu, Abhinav Gupta, and Ali Farhadi.
\newblock {AI2-THOR: An Interactive 3D Environment for Visual AI}.
\newblock \emph{arXiv}, 2017.

\bibitem[Li et~al.(2023{\natexlab{a}})Li, Li, Savarese, and Hoi]{li2023blip}
Junnan Li, Dongxu Li, Silvio Savarese, and Steven Hoi.
\newblock Blip-2: Bootstrapping language-image pre-training with frozen image encoders and large language models.
\newblock \emph{arXiv preprint arXiv:2301.12597}, 2023{\natexlab{a}}.

\bibitem[Li et~al.(2023{\natexlab{b}})Li, Yin, Li, Chen, Wang, Ren, Li, Yang, Xu, Sun, Kong, and Liu]{li2023m3it}
Lei Li, Yuwei Yin, Shicheng Li, Liang Chen, Peiyi Wang, Shuhuai Ren, Mukai Li, Yazheng Yang, Jingjing Xu, Xu~Sun, Lingpeng Kong, and Qi~Liu.
\newblock M$^3$it: A large-scale dataset towards multi-modal multilingual instruction tuning.
\newblock \emph{arXiv preprint arXiv:2306.04387}, 2023{\natexlab{b}}.

\bibitem[Li et~al.(2023{\natexlab{c}})Li, Song, Yu, Yu, Li, Huang, and Li]{li2023apibank}
Minghao Li, Feifan Song, Bowen Yu, Haiyang Yu, Zhoujun Li, Fei Huang, and Yongbin Li.
\newblock Api-bank: A benchmark for tool-augmented llms, 2023{\natexlab{c}}.

\bibitem[Li et~al.(2022)Li, Puig, Paxton, Du, Wang, Fan, Chen, Huang, Aky{\"u}rek, Anandkumar, et~al.]{li2022pre}
Shuang Li, Xavier Puig, Chris Paxton, Yilun Du, Clinton Wang, Linxi Fan, Tao Chen, De-An Huang, Ekin Aky{\"u}rek, Anima Anandkumar, et~al.
\newblock Pre-trained language models for interactive decision-making.
\newblock \emph{Advances in Neural Information Processing Systems}, 35:\penalty0 31199--31212, 2022.

\bibitem[Liang et~al.(2022)Liang, Zhang, Kwon, Yeung, and Zou]{Liang2022MindTG}
Weixin Liang, Yuhui Zhang, Yongchan Kwon, Serena Yeung, and James~Y. Zou.
\newblock Mind the gap: Understanding the modality gap in multi-modal contrastive representation learning.
\newblock \emph{ArXiv}, abs/2203.02053, 2022.
\newblock URL \url{https://api.semanticscholar.org/CorpusID:247244904}.

\bibitem[Liu et~al.(2023{\natexlab{a}})Liu, Li, Wu, and Lee]{liu2023llava}
Haotian Liu, Chunyuan Li, Qingyang Wu, and Yong~Jae Lee.
\newblock Visual instruction tuning.
\newblock 2023{\natexlab{a}}.

\bibitem[Liu et~al.(2023{\natexlab{b}})Liu, Yang, Jia, Zhang, Zhou, Dai, Yang, and Vosoughi]{liu2023training}
Ruibo Liu, Ruixin Yang, Chenyan Jia, Ge~Zhang, Denny Zhou, Andrew~M Dai, Diyi Yang, and Soroush Vosoughi.
\newblock Training socially aligned language models in simulated human society.
\newblock \emph{arXiv preprint arXiv:2305.16960}, 2023{\natexlab{b}}.

\bibitem[Lu et~al.(2023)Lu, Peng, Cheng, Galley, Chang, Wu, Zhu, and Gao]{lu2023chameleon}
Pan Lu, Baolin Peng, Hao Cheng, Michel Galley, Kai-Wei Chang, Ying~Nian Wu, Song-Chun Zhu, and Jianfeng Gao.
\newblock Chameleon: Plug-and-play compositional reasoning with large language models.
\newblock \emph{arXiv preprint arXiv:2304.09842}, 2023.

\bibitem[Marino et~al.(2019)Marino, Rastegari, Farhadi, and Mottaghi]{marino2019okvqa}
Kenneth Marino, Mohammad Rastegari, Ali Farhadi, and Roozbeh Mottaghi.
\newblock {OK-VQA:} {A} visual question answering benchmark requiring external knowledge.
\newblock In \emph{{IEEE} Conference on Computer Vision and Pattern Recognition, {CVPR} 2019, Long Beach, CA, USA, June 16-20, 2019}, pp.\  3195--3204, 2019.

\bibitem[Nakano et~al.(2022)Nakano, Hilton, Balaji, Wu, Ouyang, Kim, Hesse, Jain, Kosaraju, Saunders, Jiang, Cobbe, Eloundou, Krueger, Button, Knight, Chess, and Schulman]{nakano2022webgpt}
Reiichiro Nakano, Jacob Hilton, Suchir Balaji, Jeff Wu, Long Ouyang, Christina Kim, Christopher Hesse, Shantanu Jain, Vineet Kosaraju, William Saunders, Xu~Jiang, Karl Cobbe, Tyna Eloundou, Gretchen Krueger, Kevin Button, Matthew Knight, Benjamin Chess, and John Schulman.
\newblock Webgpt: Browser-assisted question-answering with human feedback, 2022.

\bibitem[OpenAI(2022)]{chatgpt}
OpenAI.
\newblock 2022.
\newblock URL \url{https://chat.openai.com/}.

\bibitem[OpenAI(2023{\natexlab{a}})]{GPT-4V}
OpenAI.
\newblock 2023{\natexlab{a}}.
\newblock URL \url{https://openai.com/blog/chatgpt-can-now-see-hear-and-speak}.

\bibitem[OpenAI(2023{\natexlab{b}})]{openai2023gpt4}
OpenAI.
\newblock Gpt-4 technical report, 2023{\natexlab{b}}.

\bibitem[Park et~al.(2023)Park, O'Brien, Cai, Morris, Liang, and Bernstein]{park2023generative}
Joon~Sung Park, Joseph~C O'Brien, Carrie~J Cai, Meredith~Ringel Morris, Percy Liang, and Michael~S Bernstein.
\newblock Generative agents: Interactive simulacra of human behavior.
\newblock \emph{arXiv preprint arXiv:2304.03442}, 2023.

\bibitem[Qin et~al.(2023)Qin, Hu, Lin, Chen, Ding, Cui, Zeng, Huang, Xiao, Han, et~al.]{qin2023tool}
Yujia Qin, Shengding Hu, Yankai Lin, Weize Chen, Ning Ding, Ganqu Cui, Zheni Zeng, Yufei Huang, Chaojun Xiao, Chi Han, et~al.
\newblock Tool learning with foundation models.
\newblock \emph{arXiv preprint arXiv:2304.08354}, 2023.

\bibitem[Radford et~al.(2019)Radford, Wu, Child, Luan, Amodei, Sutskever, et~al.]{radford2019language}
Alec Radford, Jeffrey Wu, Rewon Child, David Luan, Dario Amodei, Ilya Sutskever, et~al.
\newblock Language models are unsupervised multitask learners.
\newblock \emph{OpenAI blog}, 1\penalty0 (8):\penalty0 9, 2019.

\bibitem[Redmon \& Farhadi(2018)Redmon and Farhadi]{YOLO}
Joseph Redmon and Ali Farhadi.
\newblock Yolov3: An incremental improvement.
\newblock \emph{arXiv}, 2018.

\bibitem[Ren et~al.(2023{\natexlab{a}})Ren, Li, Ren, Zhao, and Sun]{ren-etal-2023-delving}
Shuhuai Ren, Lei Li, Xuancheng Ren, Guangxiang Zhao, and Xu~Sun.
\newblock Delving into the openness of {CLIP}.
\newblock In \emph{Findings of the Association for Computational Linguistics: ACL 2023}. Association for Computational Linguistics, July 2023{\natexlab{a}}.
\newblock URL \url{https://aclanthology.org/2023.findings-acl.610}.

\bibitem[Ren et~al.(2023{\natexlab{b}})Ren, Zhang, Zhu, Zhang, Zheng, Li, Smola, and Sun]{ren2023pomp}
Shuhuai Ren, Aston Zhang, Yi~Zhu, Shuai Zhang, Shuai Zheng, Mu~Li, Alex Smola, and Xu~Sun.
\newblock Prompt pre-training with twenty-thousand classes for open-vocabulary visual recognition.
\newblock \emph{arXiv preprint arXiv:2304.04704}, 2023{\natexlab{b}}.

\bibitem[Schick et~al.(2023)Schick, Dwivedi-Yu, Dessì, Raileanu, Lomeli, Zettlemoyer, Cancedda, and Scialom]{schick2023toolformer}
Timo Schick, Jane Dwivedi-Yu, Roberto Dessì, Roberta Raileanu, Maria Lomeli, Luke Zettlemoyer, Nicola Cancedda, and Thomas Scialom.
\newblock Toolformer: Language models can teach themselves to use tools, 2023.

\bibitem[Shridhar et~al.(2020)Shridhar, Thomason, Gordon, Bisk, Han, Mottaghi, Zettlemoyer, and Fox]{ALFRED20}
Mohit Shridhar, Jesse Thomason, Daniel Gordon, Yonatan Bisk, Winson Han, Roozbeh Mottaghi, Luke Zettlemoyer, and Dieter Fox.
\newblock {ALFRED: A Benchmark for Interpreting Grounded Instructions for Everyday Tasks}.
\newblock In \emph{The IEEE Conference on Computer Vision and Pattern Recognition (CVPR)}, 2020.
\newblock URL \url{https://arxiv.org/abs/1912.01734}.

\bibitem[Song et~al.(2023)Song, Xiong, Zhu, Wu, Qian, Song, Huang, Li, Wang, Yao, Tian, and Li]{song2023restgpt}
Yifan Song, Weimin Xiong, Dawei Zhu, Wenhao Wu, Han Qian, Mingbo Song, Hailiang Huang, Cheng Li, Ke~Wang, Rong Yao, Ye~Tian, and Sujian Li.
\newblock Restgpt: Connecting large language models with real-world restful apis, 2023.

\bibitem[Suhr et~al.(2017)Suhr, Lewis, Yeh, and Artzi]{Suhr2017NLVR}
Alane Suhr, Mike Lewis, James Yeh, and Yoav Artzi.
\newblock A corpus of natural language for visual reasoning.
\newblock In \emph{Proceedings of the 55th Annual Meeting of the Association for Computational Linguistics (Volume 2: Short Papers)}, pp.\  217--223, 2017.

\bibitem[Touvron et~al.(2023)Touvron, Lavril, Izacard, Martinet, Lachaux, Lacroix, Rozi{\`e}re, Goyal, Hambro, Azhar, et~al.]{touvron2023llama1}
Hugo Touvron, Thibaut Lavril, Gautier Izacard, Xavier Martinet, Marie-Anne Lachaux, Timoth{\'e}e Lacroix, Baptiste Rozi{\`e}re, Naman Goyal, Eric Hambro, Faisal Azhar, et~al.
\newblock Llama: Open and efficient foundation language models.
\newblock \emph{arXiv preprint arXiv:2302.13971}, 2023.

\bibitem[Wang et~al.(2023{\natexlab{a}})Wang, Xie, Jiang, Mandlekar, Xiao, Zhu, Fan, and Anandkumar]{wang2023voyager}
Guanzhi Wang, Yuqi Xie, Yunfan Jiang, Ajay Mandlekar, Chaowei Xiao, Yuke Zhu, Linxi Fan, and Anima Anandkumar.
\newblock Voyager: An open-ended embodied agent with large language models.
\newblock \emph{arXiv preprint arXiv:2305.16291}, 2023{\natexlab{a}}.

\bibitem[Wang et~al.(2023{\natexlab{b}})Wang, Li, Chen, Song, Lin, Cao, Liu, and Sui]{wang2023making}
Peiyi Wang, Lei Li, Liang Chen, Feifan Song, Binghuai Lin, Yunbo Cao, Tianyu Liu, and Zhifang Sui.
\newblock Making large language models better reasoners with alignment.
\newblock \emph{arXiv preprint arXiv:2309.02144}, 2023{\natexlab{b}}.

\bibitem[Wang et~al.(2023{\natexlab{c}})Wang, Li, Chen, Zhu, Lin, Cao, Liu, Liu, and Sui]{wang2023large}
Peiyi Wang, Lei Li, Liang Chen, Dawei Zhu, Binghuai Lin, Yunbo Cao, Qi~Liu, Tianyu Liu, and Zhifang Sui.
\newblock Large language models are not fair evaluators.
\newblock \emph{arXiv preprint arXiv:2305.17926}, 2023{\natexlab{c}}.

\bibitem[Wang et~al.(2023{\natexlab{d}})Wang, Cai, Liu, Ma, and Liang]{Wang2023DescribeEP}
Zihao Wang, Shaofei Cai, Anji Liu, Xiaojian Ma, and Yitao Liang.
\newblock Describe, explain, plan and select: Interactive planning with large language models enables open-world multi-task agents.
\newblock \emph{ArXiv}, abs/2302.01560, 2023{\natexlab{d}}.
\newblock URL \url{https://api.semanticscholar.org/CorpusID:256598146}.

\bibitem[Wayve(2023)]{lingo}
Wayve.
\newblock Lingo.
\newblock 2023.
\newblock URL \url{https://wayve.ai/thinking/lingo-natural-language-autonomous-driving/}.

\bibitem[Wei et~al.(2022)Wei, Tay, Bommasani, Raffel, Zoph, Borgeaud, Yogatama, Bosma, Zhou, Metzler, Chi, Hashimoto, Vinyals, Liang, Dean, and Fedus]{wei2022emergent}
Jason Wei, Yi~Tay, Rishi Bommasani, Colin Raffel, Barret Zoph, Sebastian Borgeaud, Dani Yogatama, Maarten Bosma, Denny Zhou, Donald Metzler, Ed~H. Chi, Tatsunori Hashimoto, Oriol Vinyals, Percy Liang, Jeff Dean, and William Fedus.
\newblock Emergent abilities of large language models, 2022.

\bibitem[Wei et~al.(2023)Wei, Wang, Schuurmans, Bosma, Ichter, Xia, Chi, Le, and Zhou]{wei2023chainofthought}
Jason Wei, Xuezhi Wang, Dale Schuurmans, Maarten Bosma, Brian Ichter, Fei Xia, Ed~Chi, Quoc Le, and Denny Zhou.
\newblock Chain-of-thought prompting elicits reasoning in large language models, 2023.

\bibitem[Wu et~al.(2023)Wu, Yin, Qi, Wang, Tang, and Duan]{Wu2023VisualCT}
Chenfei Wu, Sheng-Kai Yin, Weizhen Qi, Xiaodong Wang, Zecheng Tang, and Nan Duan.
\newblock Visual chatgpt: Talking, drawing and editing with visual foundation models.
\newblock \emph{ArXiv}, abs/2303.04671, 2023.
\newblock URL \url{https://api.semanticscholar.org/CorpusID:257404891}.

\bibitem[Xi et~al.(2023)Xi, Chen, Guo, He, Ding, Hong, Zhang, Wang, Jin, Zhou, Zheng, Fan, Wang, Xiong, Zhou, Wang, Jiang, Zou, Liu, Yin, Dou, Weng, Cheng, Zhang, Qin, Zheng, Qiu, Huang, and Gui]{xi2023rise}
Zhiheng Xi, Wenxiang Chen, Xin Guo, Wei He, Yiwen Ding, Boyang Hong, Ming Zhang, Junzhe Wang, Senjie Jin, Enyu Zhou, Rui Zheng, Xiaoran Fan, Xiao Wang, Limao Xiong, Yuhao Zhou, Weiran Wang, Changhao Jiang, Yicheng Zou, Xiangyang Liu, Zhangyue Yin, Shihan Dou, Rongxiang Weng, Wensen Cheng, Qi~Zhang, Wenjuan Qin, Yongyan Zheng, Xipeng Qiu, Xuanjing Huang, and Tao Gui.
\newblock The rise and potential of large language model based agents: A survey, 2023.

\bibitem[Yang et~al.(2023)Yang, Li, Wang, Lin, Azarnasab, Ahmed, Liu, Liu, Zeng, and Wang]{Yang2023MMREACTPC}
Zhengyuan Yang, Linjie Li, Jianfeng Wang, Kevin Lin, Ehsan Azarnasab, Faisal Ahmed, Zicheng Liu, Ce~Liu, Michael Zeng, and Lijuan Wang.
\newblock Mm-react: Prompting chatgpt for multimodal reasoning and action.
\newblock \emph{ArXiv}, abs/2303.11381, 2023.
\newblock URL \url{https://api.semanticscholar.org/CorpusID:257637012}.

\bibitem[Yao et~al.(2022)Yao, Zhao, Yu, Du, Shafran, Narasimhan, and Cao]{yao2022react}
Shunyu Yao, Jeffrey Zhao, Dian Yu, Nan Du, Izhak Shafran, Karthik~R Narasimhan, and Yuan Cao.
\newblock React: Synergizing reasoning and acting in language models.
\newblock In \emph{The Eleventh International Conference on Learning Representations}, 2022.

\bibitem[Yao et~al.(2023)Yao, Heinecke, Niebles, Liu, Feng, Xue, Murthy, Chen, Zhang, Arpit, et~al.]{yao2023retroformer}
Weiran Yao, Shelby Heinecke, Juan~Carlos Niebles, Zhiwei Liu, Yihao Feng, Le~Xue, Rithesh Murthy, Zeyuan Chen, Jianguo Zhang, Devansh Arpit, et~al.
\newblock Retroformer: Retrospective large language agents with policy gradient optimization.
\newblock \emph{arXiv preprint arXiv:2308.02151}, 2023.

\bibitem[Yuan et~al.(2023)Yuan, Zhang, Wang, Xie, Cai, Dong, and Lu]{yuan2023plan4mc}
Haoqi Yuan, Chi Zhang, Hongcheng Wang, Feiyang Xie, Penglin Cai, Hao Dong, and Zongqing Lu.
\newblock {Plan4MC}: Skill reinforcement learning and planning for open-world {Minecraft} tasks.
\newblock \emph{arXiv preprint arXiv:2303.16563}, 2023.

\bibitem[Zhao et~al.(2023)Zhao, Cai, Si, Ma, An, Chen, Liu, Wang, Han, and Chang]{zhao2023mmicl}
Haozhe Zhao, Zefan Cai, Shuzheng Si, Xiaojian Ma, Kaikai An, Liang Chen, Zixuan Liu, Sheng Wang, Wenjuan Han, and Baobao Chang.
\newblock Mmicl: Empowering vision-language model with multi-modal in-context learning.
\newblock \emph{arXiv preprint arXiv:2309.07915}, 2023.

\bibitem[Zheng et~al.(2023)Zheng, Chiang, Sheng, Zhuang, Wu, Zhuang, Lin, Li, Li, Xing, Zhang, Gonzalez, and Stoica]{zheng2023judging}
Lianmin Zheng, Wei-Lin Chiang, Ying Sheng, Siyuan Zhuang, Zhanghao Wu, Yonghao Zhuang, Zi~Lin, Zhuohan Li, Dacheng Li, Eric.~P Xing, Hao Zhang, Joseph~E. Gonzalez, and Ion Stoica.
\newblock Judging llm-as-a-judge with mt-bench and chatbot arena, 2023.

\bibitem[Zhu et~al.(2023{\natexlab{a}})Zhu, Chen, Shen, Li, and Elhoseiny]{zhu2023minigpt}
Deyao Zhu, Jun Chen, Xiaoqian Shen, Xiang Li, and Mohamed Elhoseiny.
\newblock Minigpt-4: Enhancing vision-language understanding with advanced large language models.
\newblock \emph{arXiv preprint arXiv:2304.10592}, 2023{\natexlab{a}}.

\bibitem[Zhu et~al.(2023{\natexlab{b}})Zhu, Chen, Tian, Tao, Su, Yang, Huang, Li, Lu, Wang, Qiao, Zhang, and Dai]{zhu2023ghost}
Xizhou Zhu, Yuntao Chen, Hao Tian, Chenxin Tao, Weijie Su, Chenyu Yang, Gao Huang, Bin Li, Lewei Lu, Xiaogang Wang, Yu~Qiao, Zhaoxiang Zhang, and Jifeng Dai.
\newblock Ghost in the minecraft: Generally capable agents for open-world environments via large language models with text-based knowledge and memory.
\newblock \emph{arXiv preprint arXiv:2305.17144}, 2023{\natexlab{b}}.

\bibitem[Zhu et~al.(2016)Zhu, Liang, Zhang, Huang, Li, and Hu]{Zhe_2016_CVPR_traffic_sign}
Zhe Zhu, Dun Liang, Songhai Zhang, Xiaolei Huang, Baoli Li, and Shimin Hu.
\newblock Traffic-sign detection and classification in the wild.
\newblock In \emph{The IEEE Conference on Computer Vision and Pattern Recognition (CVPR)}, 2016.

\end{thebibliography}
\bibliographystyle{iclr2024_conference}

\appendix
\newpage

\section{Examples of PCA-EVAL}
\label{app:example_pca_eval}
\subsection{Data Distribution}
The PCA-EVAL benchmark data distribution across various domains is outlined in Table~\ref{tab:data-distribution}.

For the Autonomous Driving domain, instances are grouped by their respective task types. In the Domestic Robot domain, instances are grouped by their locations. In the Open-World Game domain, instances are grouped by the tasks they aim to accomplish.




\begin{table}[htbp]
\centering
\caption{Data Distribution in the PCA-EVAL Benchmark}
\label{tab:data-distribution}
\begin{tabular}{lccc}
\toprule
\textbf{Domain} & \textbf{Task Type/Location} & \textbf{Instances} &  \\
\midrule
Autonomous Driving & Traffic Sign Detection & 44  \\
& Car Detection & 33  \\
& Human Detection & 30  \\
& Weather Detection & 9  \\
& Road Detection & 3  \\
& Character Recognition & 13  \\
\midrule
Domestic Robot  & Living Room &31 \\
 & Dining Room& 11 \\
 & Bedroom& 6 \\
  & Bathroom &3 \\
  & Kitchen &37 \\
  & Corridor &12 \\
\midrule
Open-World Game & Find Objects & 52 \\
 & Kill Enemies & 6 \\
 & Craft Items & 32 \\
 & Place Blocks & 7 \\
 & Interact with Creatures & 3 \\
\bottomrule
\end{tabular}
\end{table}

\newpage
\subsection{PCA-EVAL Examples}

\begin{figure}[h]
    \centering
    \includegraphics[width=0.9\textwidth]{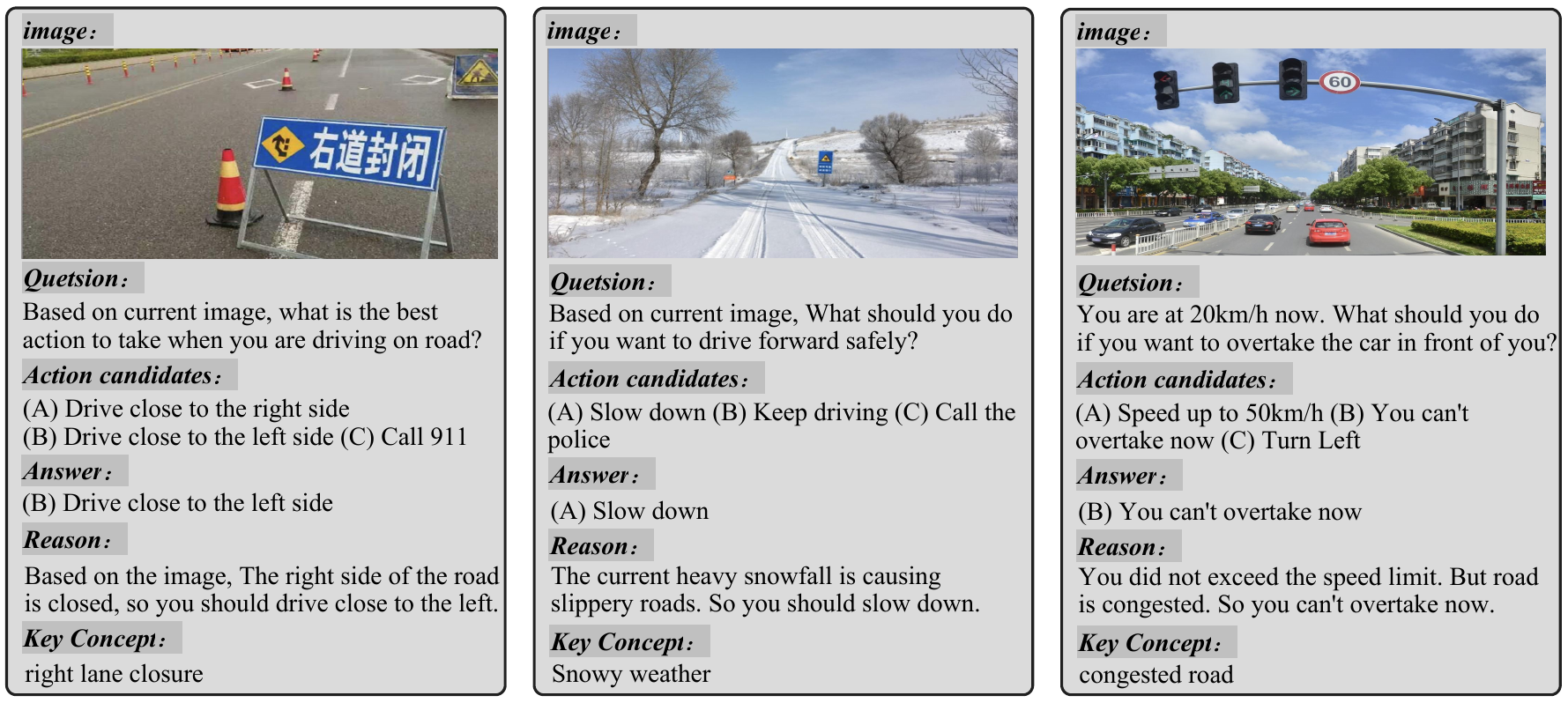}
    \caption{Three examples of PCA-EVAL in the autonomous driving domain.
    }
    \label{fig:pca-examples-traffic}
\end{figure}

\begin{figure}[h]
    \centering
    \includegraphics[width=0.90\textwidth]{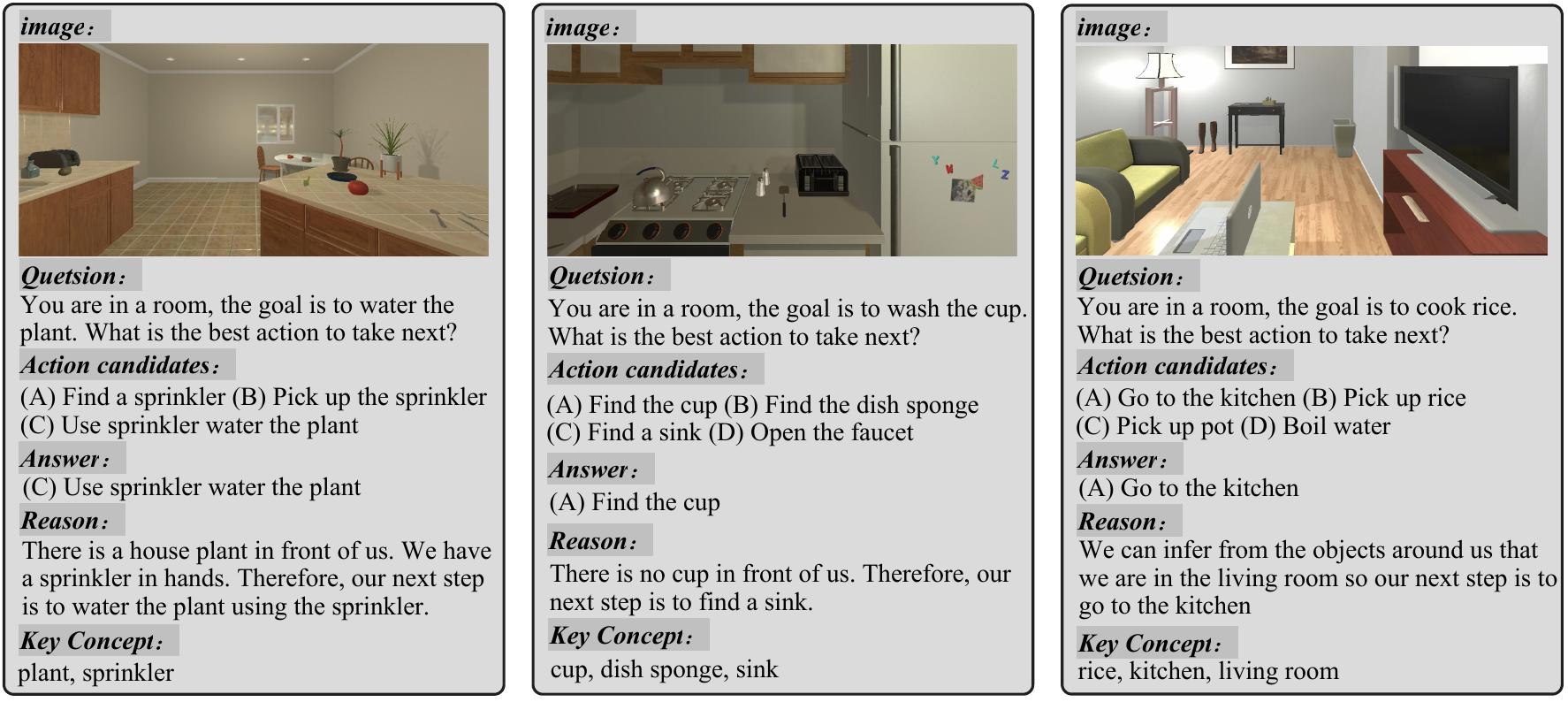}
    \caption{Three examples of PCA-EVAL in the domestic robot domain.
    }
    \label{fig:pca-examples-domestic}
\end{figure}

\begin{figure}[h]
    \centering
    \includegraphics[width=0.83\textwidth]{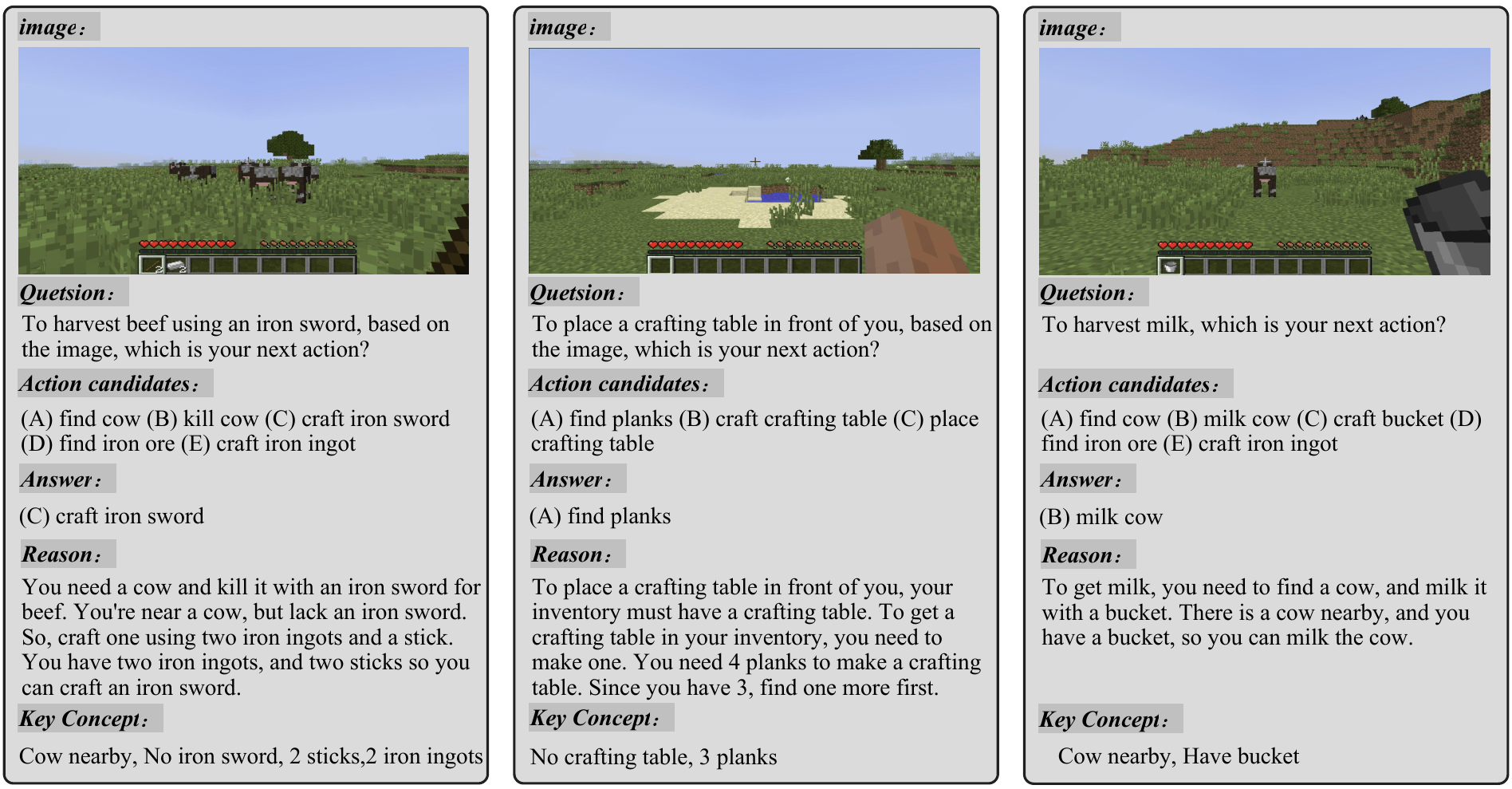}
    \caption{Three examples of PCA-EVAL in the open-world game domain.
    }
    \label{fig:pca-examples-game}
\end{figure}

We list three examples of each domain from PCA-EVAL, as shown in Figure~\ref{fig:pca-examples-traffic}, Figure~\ref{fig:pca-examples-domestic} and Figure~\ref{fig:pca-examples-game}.

\newpage

\section{API Description and Implementation of HOLMES}
\label{app:api_des}
\paragraph{Traffic Domain.} Below is the API description for the traffic domain.
\begin{lstlisting}[language=Python]
# API Description for Traffic Domain:
def detect_traffic_sign():
    """
    Detects traffic signs in the image.
    :return: list of detected traffic signs and coordinates, e.g. ['stop','max speed limit']
    """
    pass

def object_detection():
    """
    Detects objects in the image.
    :return: dict of detected objects and number of the objects, e.g. {'car':10, 'person':1}
    """
    pass

def ocr():
    """
    Performs OCR on the image.
    :return: list of detected text, e.g. ['Changjiang road', 'Right lane closure']
    """
    pass

def image_caption():
    """
    Generates a caption for the image.
    :return: caption, e.g. 'A red car driving down the street'
    """
    pass

def weather_detection():
    """
    Detect current weather.
    :return: weather, e.g. 'rainy' or 'clear'
    """
    pass
\end{lstlisting}
\textbullet\ \textit{detect\_traffic\_sign()}: The detection of road traffic signs model utilize YOLO~\citep{YOLO} which trained on the Tsinghua-Tencent 100K dataset~\citep{Zhe_2016_CVPR_traffic_sign}. TT100K comprises 100,000 images encompassing 30,000 instances of traffic signs. The end-to-end YOLO enables simultaneous detection and classification of traffic signs.

\textbullet\ \textit{object\_detection()}: Objects demanding attention during vehicle operation primarily encompass cars, pedestrians, and bicycles. A surfeit of vehicles can lead to traffic congestion, while the presence of pedestrians or bicycles ahead necessitates cars to decelerate and proceed cautiously. Hence, the \textit{object\_detection()} API predominantly identifies three key object categories: cars, pedestrians, and bicycles. We utilize PMOP~\citep{ren2023pomp}, a model trained on vision-language models through the prompt pre-training method, which enables the detection and counting of the three mentioned objectives by modifying specific class names.

\textbullet\ \textit{ocr()}: We employ PaddleOCR\footnote{\url{https://github.com/PaddlePaddle/PaddleOCR/tree/release/2.7}} to extract textual information from images, providing crucial road data for real-time navigation.

\textbullet\ \textit{image\_caption()}: To initially streamline the road information within the image, we employ the BLIP2-flan-t5-xl\footnote{\url{https://huggingface.co/Salesforce/blip2-flan-t5-xl}} to generate an initial caption for the picture. This caption, derived from basic image data, is then utilized as input for the model to facilitate decision-making.

\textbullet\ \textit{weather\_detection()}: Weather detection leverages a pre-trained ResNet50 model\footnote{\url{https://github.com/mengxianglong123/weather-recognition}}, derived from a dataset of more than 70,000 weather records. This model extracts weather information from provided images to inform decision-making.

\paragraph{Domestic Robot Domain.} Below is the API description for the Domestic Robot domain.
\begin{lstlisting}[language=Python]
#API Description for Domestic Robot Domain
def object\_detection():
    """
    Detects objects in current view, which you don't need do find.
    :return: list of detected objects, e.g. ['chair','table']
    """
    pass

def list_items_in_hands():
    """
    Lists items in your hand, which you don't need to pick up
    :return: list of items in hand, e.g. ['coffee cup','milk']
    """
    pass
\end{lstlisting}

\paragraph{Game Domain.} Below is the API description for the Game domain (Minedojo).
\begin{lstlisting}[language=Python]
#API Description for Game Domain
def list_nearby_mobs_in_minecraft():
    """
    Lists nearby mobs in Minecraft.
    :return: list of nearby mobs, e.g. ['creeper', 'pig']
    """
    pass

def list_inventory_information():
    """
    Lists inventory information of the player in Minecraft.
    :return: list of inventory information with number, e.g. [('diamond', 64), ('iron', 32)]
    """
    pass
\end{lstlisting}

Note that within the Domestic Robot Domain and Game Domain, APIs can be directly accessed within the virtual environment, allowing for the perception of the surrounding objects and the current picture context.

\newpage
\section{Automatic Evaluation}
\label{app:ae}

\begin{table*}[ht!]

\begin{tcolorbox}

[Question]: \{question\}

[Action Choices]: \{actions\}

[Agent Answer]: \{model\_output\}

[Correct Action]: \{true\_action\}

[Key Concepts]: \{key\_concept\}

[Reference Reasoning Process]: \{reason\}

[System]

We would like you to access the agent's performance in the multimodal reasoning task about {domain}.
In this task, the agent is given an image, a [Question], and several candidate [Action Choices], and is asked to give an [Agent Answer] for the [Question].
The [Agent Answer] encapsulates the agent's perception of the image's [Key Concepts], the agent's cognition reasoning process and the final selected action.

We request you to give three types of scores for the agent's [Agent Answer] in comparison to the given [Key Concepts], [Reference Reasoning Process] and [Correct Action]:

1. action score: If the selected action in the [Agent Answer] matches that of the [Correct Action], the action score is 1; otherwise, it is 0.

2. perception score: This score evaluates the model's capability to perceive and interpret observations. It is contingent on whether the [Agent Answer] includes any of the [Key Concepts] of the instance. If it accurately describes any one of the [Key Concepts], the score is 1; otherwise, it is 0.

3. cognition score: This score gauges the model's ability to reason, comprehend, and make informed decisions based on perceived input data and world knowledge. If the reasoning process in the [Agent Answer] aligns with the [Reference Reasoning Process], the score is 1; otherwise, it is 0.

Please note that there are only scores of 0 and 1.

You should carefully compare the [Agent Answer] with the [Correct Action], [Key Concepts] and [Reference Reasoning Process] to give your assessment.

You need first to give your assessment evidence and then the scores. 

Your output MUST contain 6 lines with the following format:

action assessment evidence: (assessment evidence here)

action score: (score here)

perception assessment evidence: (assessment evidence here)

perception score: (score here)

cognition assessment evidence: (assessment evidence here)

cognition score: (score here)

\end{tcolorbox}
\caption{We utilize the template to query GPT-4, aiming to evaluate its responses and assign scores for perception, cognition, and action. By feeding both the agent's output and the ground truth answer to GPT-4, based on this template, we can then extract the three distinct scores from the conclusion of GPT-4's response.}
\label{tab:gpt4-eval}
\end{table*}




\end{document}